\documentclass[11pt,a4paper]{article}
\usepackage{authblk}
\usepackage[hyperref]{acl2021}
\usepackage{times}
\usepackage{latexsym}
\usepackage{amsmath}
\usepackage{booktabs}
\usepackage{amssymb}
\usepackage{colortbl}
\usepackage[colorinlistoftodos]{todonotes}
\usepackage{multirow}
\usepackage{hyperref}

\usepackage{microtype}

\aclfinalcopy 
\def\aclpaperid{67} 

\title{OCHADAI-KYOTO at SemEval-2021 Task 1:  Enhancing Model Generalization and Robustness for Lexical Complexity Prediction}

\author[1]{\bf{Yuki Taya}}
\author[1]{\bf{Lis Kanashiro Pereira}}
\author[2]{\bf{Fei Cheng}}
\author[1]{\bf{Ichiro Kobayashi}}
\affil[1]{Ochanomizu University, Japan}
\affil[2]{Kyoto University, Japan}

\date{}
\begin{document}

\maketitle

\begin{abstract}
We propose an ensemble model for predicting the lexical complexity of words and multiword expressions (MWEs). The model receives as input a sentence with a target word or MWE and outputs its complexity score.
Given that a key challenge with this task is the limited size of annotated data, our model relies on pretrained contextual representations from different state-of-the-art transformer-based language models (i.e., BERT and RoBERTa), and on a variety of training methods for further enhancing model generalization and robustness: multi-step fine-tuning and multi-task learning, and adversarial training. Additionally, we propose to enrich contextual representations by adding hand-crafted features during training. Our model achieved competitive results and ranked among the top-10 systems in both sub-tasks.
\end{abstract}

\section{Introduction}
Predicting the difficulty of a word in a given context is useful in many natural language processing (NLP) applications such as lexical simplification. Previous efforts \cite{paetzold2016semeval,yimam2018report,zampieri2017complex} have focused on framing this as a binary classification task, which might not be ideal, since a word close to the decision
boundary is assumed to be just as complex as one further
away \cite{shardlow2020complex}. To alleviate this issue, SemEval-2021 Task 1 \cite{shardlow2021semeval} formulates this task as a regression task, where a model should predict the complexity value of words (Subtask 1) and MWEs (Subtask 2) in context. 

This paper describes the system developed by the Ochadai-Kyoto team for SemEval-2021 Task 1. Given that a key challenge in this task is the limited size of annotated data, we follow best practices from recent work on enhancing model generalization and robustness, and propose a model ensemble that leverages pretrained representations (i.e. BERT and RoBERTa), multi-step fine-tuning, multi-task learning and adversarial training. Additionally, we propose to enrich contextual representations by incorporating hand-crafted features during training. Our model ranked 7th out of 54 participating teams on Subtask 1, and 8th out of 37 teams on Subtask 2, obtaining Pearson correlation scores of 0.7772 and 0.8438, respectively.

\section{Task Description}

SemEval-2021 Task 1 provides participants with an augmented version of the CompLex dataset \cite{shardlow2020complex}, a multi-domain English dataset with sentences containing words and MWEs annotated on a continuum scale of complexity, in the range of [0,1]. Easier words and MWEs are assigned lower complexity scores, while the more challenging ones are assigned higher scores. This corpus contains a balanced number of sentences from three different domains: Bible \cite{christodouloupoulos2015massively}, Europarl \cite{koehn2005europarl} and Biomedical \cite{bada2012concept}. The task is to predict the complexity value of single words (Subtask 1) and MWEs (Subtask 2) in context. The statistics of the corpus are presented in Table~\ref{data1-table}. Our team participated in both subtasks, and the next section outlines the overview of our model.

\begin{table}
\small
\centering
\begin{tabular}{l|l|c|c|c} \hline
\toprule
 \bf{Task} & \bf{Domain}& \bf{Train} & \bf{Trial} & \bf{Test} \\ \hline
 \multirow{4}{1.8cm}{Subtask 1 \small (single-word)} &Europarl   &    2512    &     143    &   345     \\
&Biomed     &    2576    &     135    &   289     \\
&Bible      &    2574    &     143    &   283     \\ 
&All        &    7662    &     421    &   917     \\ \hline
\multirow{4}{1.8cm}{Subtask 2 \small (MWE)}&Europarl   &     498    &     37     &     65    \\
&Biomed     &     514    &     33     &     53    \\
&Bible      &     505    &     29     &     66    \\ 
&All        &    1517    &     99     &    184    \\
\bottomrule
\end{tabular}
\caption{\label{data1-table} Summary of the Complex dataset.}
\end{table}

\section{System Overview}
\label{sec:model}
We focus on exploring different training techniques using BERT and RoBERTa, given their superior performance on a wide range of NLP tasks. Each text encoder and training method used in our model are detailed below.

\subsection{Text Encoders}
\noindent\textbf{BERT} \cite{devlin-etal-2019-bert}: We use the BERT\textsubscript{BASE} model released by the authors. It consists of 12 transformer layers, 12 self-attention heads per layer, and a hidden size of 768.

\noindent\textbf{RoBERTa} \cite{liu2019roberta}: We use both the RoBERTa\textsubscript{BASE} and RoBERTa\textsubscript{LARGE} models released by the authors. Similar to BERT, RoBERTa\textsubscript{BASE} consists
of 12 transformer layers, 12 self-attention
heads per layer, and a hidden size of 768. RoBERTa\textsubscript{LARGE} consists
of 24 transformer layers, 16 self-attention
heads per layer, and a hidden size of 1024.

\subsection{Training Procedures}
\noindent\textbf{Standard fine-tuning}: 
This is the standard fine-tuning procedure where we fine-tune BERT and RoBERTa on each subtask-specific data.

\noindent\textbf{Feature-enriched fine-tuning (FEAT)}: During training, we enrich BERT and RoBERTa representations with word frequency information of the target word or MWE. We compute the log frequency values using the Wiki40B corpus \cite{guo2020wiki}. For MWEs, we compute the log of the average of the frequency of each component word. After applying the min-max normalization to this feature, we concatenate
it to the CLS token vector obtained from the last layer of BERT and RoBERTa. 

\noindent\textbf{Multi-step fine-tuning (MSFT)}: Multi-step fine-tuning works by performing a second stage of pretraining with data-rich related supervised tasks. It has been shown to improve model robustness and performance, especially for data-constrained scenarios \cite{phang2018sentence,camburu2019surprisingly}. Due to the limited size of the data provided for Subtask 2, we first fine-tune BERT and RoBERTa on the Subtask 1 dataset. This model's parameters are further refined by fine-tuning on the Subtask 2 dataset.

\noindent\textbf{Multi-task learning (MTL)}: Multi-task learning is an effective training paradigm to promote model generalization ability and performance \cite{caruana1997multitask,liu2015representation,liu2019multi, ruder2017overview,collobert2011natural}. It works by leveraging data from many (related) tasks. In our experiments, we use the MT-DNN framework \cite{liu2019multi,liu2020mtmtdnn}, which incorporates BERT and RoBERTa as the shared text encoding layers (shared across all tasks), while the top layers are task-specific. We used the pre-trained BERT and RoBERTa models to initialize its shared layers and refined them via MTL on both subtasks (i.e. Subtask 1 and Subtask 2).

\noindent\textbf{Adversarial training (ADV)}: Adversarial training has proven effective in improving model generalization and robustness in computer vision \cite{madry2017pgd,goodfellow2014explaining} and more recently in NLP \cite{zhu2019freelb,jiang2019smart,cheng2019robust,liu2020alum,pereira2020alic}. It works by augmenting the input with a small perturbation that maximizes the adversarial loss:
\begin{equation}
\min_{\theta} \mathbb{E}_{(x, y)\sim D}[\max_{ \delta} l(f(x + \delta; \theta), y)]
\label{eq:adv}
\end{equation}
where the inner maximization can be solved by projected gradient descent \cite{madry2017pgd}. Recently, adversarial training has been successfully applied to NLP as well \cite{zhu2019freelb,jiang2019smart,pereira2020alic}. In our experiments, we use SMART \cite{jiang2019smart}, which instead regularizes the standard training objective using {\em virtual adversarial training} \cite{miyato2018virtual}:
\begin{equation}
\begin{aligned}
  \min_{\theta} \mathbb{E}_{(x, y)\sim D}[l(f(x; \theta), y) + \\ \alpha  \max_{
\delta} l(f(x+\delta; \theta), f(x; \theta))]
\end{aligned}
\vspace{-1mm}
\label{eq:alum}
\end{equation}
Effectively, the adversarial term encourages smoothness in the input neighborhood, and $\alpha$ is a hyperparameter that controls the trade-off between standard errors and adversarial errors.

\subsection{Ensemble Model}
Ensemble of deep learning models has proven effective in improving test accuracy \cite{allen2020towards}. We built different ensemble models by taking an unweighted average of the outputs of a few independently trained models. Each single model was trained on standard fine-tuning, multi-step fine-tuning, multi-task learning, or adversarial training, using different text encoders (i.e. BERT or RoBERTa).

\section{Experiments}
\subsection{Implementation Details}
Our model implementation is based on
the MT-DNN framework \cite{liu2019multi,liu2020mtmtdnn}. We use BERT \cite{devlin-etal-2019-bert} and RoBERTa \cite{liu2019roberta} as the text encoders.
We used ADAM \cite{kingma2014adam} as our optimizer with a learning rate in the range $\in \{8 \times 10^{-6}, 9 \times 10^{-6}, 1 \times 10^{-5}\}$ and a batch size $\in \{8, 16, 32\}$. The maximum number of epochs was set to 10. A linear learning rate decay schedule with warm-up over 0.1 was used, unless stated otherwise. To avoid gradient exploding, we clipped the gradient norm within 1. All the texts were tokenized using wordpieces and were chopped to spans no longer than 512 tokens. During adversarial training, we follow \cite{jiang2019smart} and set the perturbation size to $1 \times 10^{-5}$, the step size to $ 1 \times 10^{-3}$, and  to $ 1 \times 10^{-5}$ the variance
for initializing the perturbation. The number of projected gradient steps and the $\alpha$ parameter (Equation 2) were both set to $1$.

We follow \cite{devlin-etal-2019-bert}, and set the first token as the [CLS] token when encoding the input. For Subtask 1, we separate the input sentence and the target token with the special token [SEP]. e.g. [CLS] This was the \textit{length} of Sarah’s life [SEP] \textit{length} [SEP]. For Subtask 2, such encoding led to lower performance of our system. Therefore, we consider only the target MWE when encoding the input, e.g. [CLS] \textit{financial world} [SEP].

For each subtask, we used the trial dataset released by organizers as development set (see Table 1). We select the best epoch and the best hyper-parameters using performance (measured in terms of Pearson correlation score) on this development set. We also experimented on saving the best epoch and best hyper-parameters for each domain (Bible, Biomedical and Europarl).

\subsection{Main Results}
\begin{table*}[!htb]
\begin{center}
\small
\begin{tabular}{@{\hskip3pt}l@{\hskip10pt}|>{\columncolor{cyan!10}}c>{\columncolor{cyan!20}}c >{\columncolor{cyan!30}}c|@{\hskip4pt}c@{\hskip4pt}|@{\hskip4pt}c@{\hskip4pt}|@{\hskip4pt}c@{\hskip4pt}|@{\hskip4pt}c@{\hskip4pt}|@{\hskip4pt}c@{\hskip4pt}}
\toprule
\bf Training Methods                                                             & \multicolumn{3}{c|}{\bf Ensemble} &     R    &    Rho   &    MAE   &    MSE   &     R2   \\ \midrule
\multicolumn{9}{c}{\bf  Subtask 1 (Single Word Lexical Complexity Prediction Task) }  \\ \hline
BERT\textsubscript{BASE}\textsuperscript{dev}                                       &          &          &          &   0.7794 &   0.7423 &   0.0664 &   0.0077 &   0.1898 \\ 
RoBERTA\textsubscript{BASE}\textsuperscript{dev}                                    &          &          &          &   0.8139 &   0.7498 &   0.0628 &   0.0064 &   0.4325 \\ 
RoBERTA\textsubscript{BASE}(\small FEAT)\textsuperscript{dev}                       &\checkmark&          &          &   0.8348 &   0.7579 & 0.0603 & 0.0058 & 0.6955 \\ 
RoBERTA\textsubscript{BASE}(\small FEAT)\textsuperscript{dev}\textsubscript{domain} &          &\checkmark&\checkmark& 0.8391 &   0.7640 & 0.0599 & 0.0057 & 0.6976 \\
RoBERTA\textsubscript{LARGE}\textsuperscript{dev}                                   &          &          &          &   0.8213 &   0.7629 &   0.0627 &   0.0062 &   0.5381 \\
RoBERTA\textsubscript{LARGE}(\small FEAT)\textsuperscript{dev}\textsubscript{domain}&          &          &          &   0.8218 &   0.7513 &   0.0634 &   0.0063 &   0.6025 \\ 
RoBERTA\textsubscript{LARGE}(\small MTL)\textsuperscript{dev}\textsubscript{domain} &          &\checkmark&\checkmark&0.8371 &0.7694 &   0.0609 &   0.0062 &   0.3640 \\
RoBERTA\textsubscript{LARGE}(\small ADV)\textsuperscript{dev}                       &\checkmark&          &          &   0.8328 & 0.7760 &0.0603 &   0.0059 &   0.5509 \\ 
RoBERTA\textsubscript{LARGE}(\small ADV)\textsuperscript{dev}\textsubscript{domain} &          &\checkmark&          &\bf0.8441 &\bf0.7873 &\bf0.0572 &\bf0.0054 &\bf0.7123 \\ \midrule
Ensemble 1\textsubscript{single\_word}\textsuperscript{dev}                         &$\bigcirc$&          &          &   0.8481 &   0.7825 &   0.0578 &   0.0053 &   0.7175 \\
Ensemble 2\textsubscript{single\_word}\textsuperscript{dev}                         &          &$\bigcirc$&          &\bf0.8570 &\bf0.7902 &\bf0.0553 &\bf0.0050 &\bf0.7335 \\
Ensemble 3\textsubscript{single\_word}\textsuperscript{dev}                         &          &          &$\bigcirc$&   0.8548 &   0.7816 &   0.0560 &   0.0051 &   0.7300 \\ \hline
Ensemble 1\textsubscript{single\_word}\textsuperscript{test}                        &$\bigcirc$&          &          &   0.7590 &   0.7174 &   0.0640 &   0.0069 &   0.5719 \\
Ensemble 2\textsubscript{single\_word}\textsuperscript{test}                        &          &$\bigcirc$&          &\bf0.7772 &\bf0.7313 &\bf0.0617 &\bf0.0065 &\bf0.6015 \\
Ensemble 3\textsubscript{single\_word}\textsuperscript{test}                        &          &          &$\bigcirc$&   0.7761 &   0.7244 &   0.0622 &   0.0065 &   0.6003 \\ \hline 
Top Team Result (JUST BLUE)\textsubscript{single\_word}\textsuperscript{test}*  &          &          &          &   \bf 0.7886 & \bf  0.7369 & \bf 0.0609 & \bf 0.0062 & \bf 0.6172 \\ \hline \hline
\end{tabular}

\begin{tabular}{@{\hskip3pt}l@{\hskip12pt}|>{\columncolor{orange!10}}c >{\columncolor{orange!20}}c >{\columncolor{orange!30}}c|@{\hskip4pt}c@{\hskip4pt}|@{\hskip4pt}c@{\hskip4pt}|@{\hskip4pt}c@{\hskip4pt}|@{\hskip4pt}c@{\hskip4pt}|@{\hskip4pt}c@{\hskip4pt}}
\multicolumn{9}{c}{\bf  Subtask 2 (MWE Lexical Complexity Prediction Task)}  \\  \hline
\toprule
BERT\textsubscript{BASE}\textsuperscript{dev}                                            &          &          &          &   0.7965 &   0.7856 &   0.0761 &   0.0086 &   0.3552 \\ 
BERT\textsubscript{BASE}(\small MSFT)\textsuperscript{dev}                               &\checkmark&          &          &   0.8370 &   0.8361 &\bf0.0661 &   0.0071 &   0.5276 \\ 
BERT\textsubscript{BASE}(\small MSFT)\textsuperscript{dev}\textsubscript{domain}         &          &\checkmark&\checkmark&\bf0.8498 &\bf0.8492 &   0.0669 &   0.0068 &   0.7099 \\ 
BERT\textsubscript{BASE}(\small MTL)\textsuperscript{dev}                                &\checkmark&          &          &   0.8176 &   0.8202 &   0.0725 &   0.0081 &   0.5086 \\
BERT\textsubscript{BASE}(\small MTL)\textsuperscript{dev}\textsubscript{domain}          &          &\checkmark&\checkmark&   0.8442 &   0.8323 &   0.0667 &\bf0.0067 &\bf0.7125 \\
RoBERTA\textsubscript{BASE}\textsuperscript{dev}                                         &          &          &          &   0.7689 &   0.7659 &   0.0771 &   0.0098 &   0.3767 \\ 
RoBERTA\textsubscript{LARGE}\textsuperscript{dev}                                        &          &          &          &   0.8110 &   0.8181 &   0.0737 &   0.0082 &   0.4363 \\
RoBERTA\textsubscript{LARGE}(\small MTL)\textsuperscript{dev}                            &\checkmark&          &          &   0.8176 &   0.8202 &   0.0725 &   0.0081 &   0.5086 \\
RoBERTA\textsubscript{LARGE}(\small MTL)\textsuperscript{dev}\textsubscript{domain}      &          &          &\checkmark&   0.8341 &   0.8276 &   0.0675 &   0.0075 &   0.6790 \\
RoBERTA\textsubscript{LARGE}(\small ADV)\textsuperscript{dev}                            &          &          &          &   0.8119 &   0.8019 &   0.0718 &   0.0080 &   0.4785 \\  
RoBERTA\textsubscript{LARGE}(\small ADV\&MSFT)\textsuperscript{dev}                      &          &          &          &   0.8247 &   0.8092 &   0.0685 &   0.0076 &   0.4748 \\
RoBERTA\textsubscript{LARGE}(\small ADV\&MSFT)\textsuperscript{dev}\textsubscript{domain}&          &\checkmark&          &   0.8283 &   0.8176 &   0.0676 &   0.0074 &   0.6858 \\ \midrule
Ensemble 1\textsubscript{MWE}\textsuperscript{dev}                                       &$\bigcirc$&          &          &   0.8461 &   0.8441 &   0.0672 &   0.0068 &   0.7080 \\
Ensemble 2\textsubscript{MWE}\textsuperscript{dev}                                       &          &$\bigcirc$&          &   0.8543 &   0.8444 &   0.0642 &\bf0.0064 &\bf0.7270 \\
Ensemble 3\textsubscript{MWE}\textsuperscript{dev}                                       &          &          &$\bigcirc$&\bf0.8571 &\bf0.8509 &\bf0.0640 &\bf0.0064 &   0.7267 \\ \hline
Ensemble 1\textsubscript{MWE}\textsuperscript{test}                                      &$\bigcirc$&          &          &\bf0.8438 &\bf0.8285 &\bf0.0660 &\bf0.0070 &\bf0.7103 \\
Ensemble 2\textsubscript{MWE}\textsuperscript{test}                                      &          &$\bigcirc$&          &   0.8376 &   0.8231 &   0.0682 &   0.0076 &   0.6840 \\
Ensemble 3\textsubscript{MWE}\textsuperscript{test}                                      &          &          &$\bigcirc$&   0.8312 &   0.8157 &   0.0708 &   0.0080 &   0.6686 \\ \hline
Top Team Result (DeepBlueAI)\textsubscript{single\_word}\textsuperscript{test}*          &          &          &          & \bf  0.8612 &  \bf  0.8526 &   \bf 0.0616 & \bf  0.0063 & \bf  0.7389 \\ 
\bottomrule  
\end{tabular}
\end{center}
\caption{Comparison of different text encoders and different training methods on the single word lexical complexity prediction task (Subtask 1) and on the MWE lexical complexity prediction task (Subtask 2). Best results for single and ensemble models are highlighted in \textbf{bold}. * indicates results obtained from the Task's official leaderboard: (https://competitions.codalab.org/competitions/27420\#results). \checkmark indicates each single model that was used in the ensemble, indicated in each column by $\bigcirc$.}
\label{tab:result-dev-single-multi}
\end{table*}

Submitted systems were evaluated on five metrics: Pearson correlation (R), Spearman correlation (Rho), Mean Absolute Error (MAE), Mean Squared Error (MSE), and R-squared (R2). The systems were ranked from highest Pearson correlation score to lowest. We built several models that use different text encoders and different training methods, as described in Section \ref{sec:model}. See \autoref{tab:result-dev-single-multi} for the results.

First, we observe that ensembling different single models yield better performance on both tasks. Furthermore, models that use feature-enriched representations, multi-task learning, multi-step fine-tuning and adversarial training surpass models that use the standard fine-tuning approach. We detail next the results for each subtask.

For Subtask 1, the single models that used RoBERTa, adversarial training, multi-task learning and feature-enriched representations performed best on the development set. Moreover, saving the best epoch and hyper-parameters for each domain performed better than saving the best epoch and hyper-parameters without domain distinction. Among the single models, the model that performed best on the development set was the model that uses RoBERTa\textsubscript{LARGE} and adversarial training (RoBERTa\textsubscript{LARGE}(ADV)\textsubscript{domain} model, with a Pearson score of 0.8441). The second best single model was the model that uses RoBERTa\textsubscript{BASE} and feature-enriched contextual representations (RoBERTa\textsubscript{BASE}(FEAT)\textsubscript{domain} model, with a Pearson score of 0.8391). The third best single model was the model that uses RoBERTa\textsubscript{LARGE} and multi-task learning (RoBERTa\textsubscript{LARGE}(MTL)\textsubscript{domain} model, with a Pearson score of 0.8371). Thus, we ensemble these three single models in different ways when making our submissions. The ensemble model that performed best on the test set (Ensemble 2\textsubscript{single\_word})  was the model that combined feature-enriched contextual representations (RoBERTa\textsubscript{BASE}(FEAT)\textsubscript{domain}), adversarial training (RoBERTa\textsubscript{LARGE}(ADV)\textsubscript{domain}), and multi-task learning (RoBERTa\textsubscript{LARGE}(MTL)\textsubscript{domain}). This ensemble model obtained development and test set Pearson scores of 0.8570 and 0.7772, respectively.

For Subtask 2, the single models that use BERT\textsubscript{BASE} outperformed models that use RoBERTa, on the development set. Moreover, we noted that using the Subtask 1 dataset as auxiliary dataset by performing multi-step fine-tuning and multi-task learning greatly help to improve the performance. For instance, the BERT\textsubscript{BASE}(MSFT) outperformed the BERT\textsubscript{BASE} model by 0.0405 Pearson correlation points (0.7965 vs 0.8370). The ensemble model that performed best on the test set (Ensemble 1\textsubscript{MWE}) was the model that combined multi-step fine-tuning and multi-task learning using BERT, i.e. BERT\textsubscript{BASE} (MSFT) and BERT\textsubscript{BASE}(MTL) models, respectively, and multi-task learning using RoBERTa (RoBERTA\textsubscript{LARGE}(MTL) model). This ensemble model obtained development and test set Pearson scores of 0.8461 and 0.8438, respectively. Different from Subtask 1, we observe that saving the best epoch and hyper-parameters for each domain on the development set performed worse than saving the best epoch and hyper-parameters without domain distinction. We hypothesize that, due to the small size of the data provided for Subtask 2,  saving the best epoch and hyper-parameters without domain distinction might avoid overfitting.

\section{Analysis}
\begin{table*}[!htb]
\begin{center}
\small
\begin{tabular}{@{\hskip0pt}c@{\hskip0pt}|@{\hskip0pt}c@{\hskip0pt}|@{\hskip2pt}c@{\hskip2pt}|@{\hskip2pt}c@{\hskip2pt}|@{\hskip2pt}c@{\hskip2pt}}
\toprule

\bf{Domain}& \bf{Sentence}& \bf{Target} & \bf{Prediction} & \bf{Label} \\ \hline
\multicolumn{5}{c}{\bf Sub-task 1}  \\ \hline
Europarl &\begin{tabular}{c}
The Swedish Presidency aims to maintain the debate on \\
animal welfare and good animal \textit{husbandry}.
\end{tabular}
&\textit{husbandry} & 0.3270 & 0.53143\\\hline
Biomed & \begin{tabular}{c}
We adopted the same strategy to investigate the relative contribution of \\
the 129 \textit{Chromosome} 1 segment and the Apcs gene to each disease trait.
\end{tabular}
& \textit{Chromosome} & 0.4865 & 0.2237 \\\hline
Bible & God has gone up with a \textit{shout}, Yahweh with the sound of a trumpet. 
& \textit{shout} &\bf{0.2032}& \bf{0.2031} \\ \hline \hline
\multicolumn{5}{c}{\bf Sub-task 2}  \\ \hline
Biomed &\begin{tabular}{c}
These studies strongly suggest that the hsp family \\
of proteins has \textit{other functions} in addition to protecting \\
proteins and cells during stress.
\end{tabular}
& \textit{other functions} & 0.2564 & 0.4167 \\ \hline
Europarl & \begin{tabular}{c}
What plans does the Commission have to introduce \\ 
\textit{eco labelling} of 'sustainable' palm oils?
\end{tabular}
& \textit{eco labelling} & 0.5277 & 0.3553 \\\hline
Bible & In the \textit{dry season}, they vanish. 
& \textit{dry season} & \bf{0.2832} & \bf{0.2857} \\ 
\bottomrule
\end{tabular}
\caption{\label{tab:error_samples} Examples of successful and poor predictions on the test set by the best ensemble models submitted for each subtask (Ensemble 2\textsubscript{single\_word} and Ensemble 1\textsubscript{multiword} models). Successful predictions are highlighted in \textbf{bold}.}
\end{center}
\end{table*}

\begin{figure}[h!]
\centering
\includegraphics[scale=0.32]{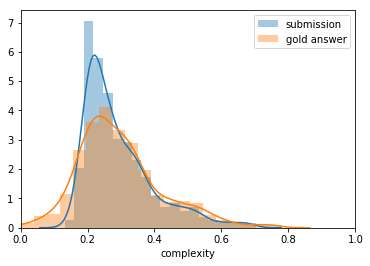}
\includegraphics[scale=0.145]{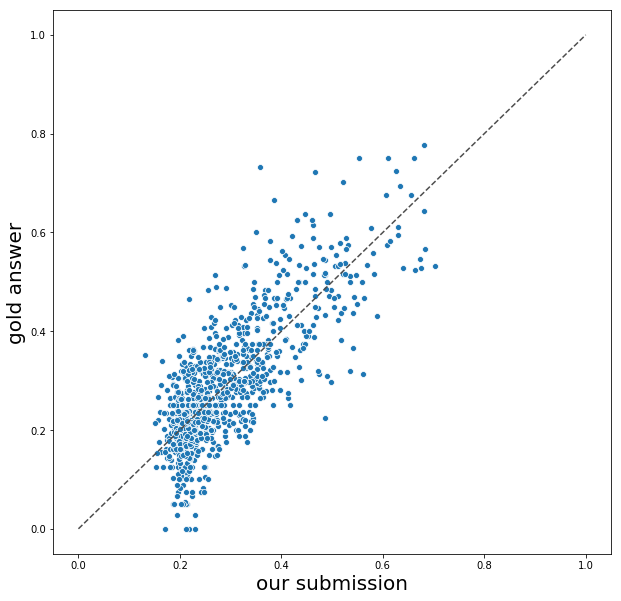}
\includegraphics[scale=0.32]{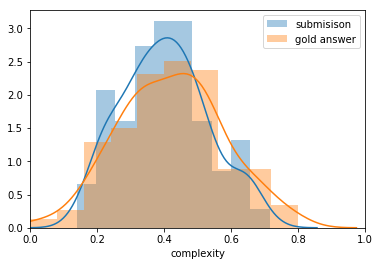}
\includegraphics[scale=0.145]{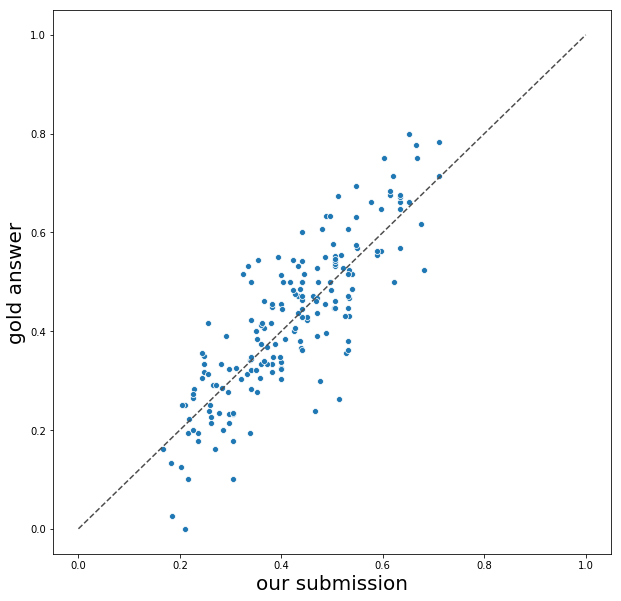}
\caption{Comparison between the Ensemble 2\textsubscript{single\_word} and Ensemble 1\textsubscript{MWE} models' predictions submitted for Sub-task 1 (top) and Sub-task 2 (bottom), respectively, and the gold answers. On the left, we show the distribution of the correct complexity score and our submission. On the right, we show a scatter plot where the x-axis corresponds to our model's predictions and the y-axis corresponds to  the gold answers.}
\vspace{-4mm}
\label{fig:single_multi_graphs}
\end{figure}

We briefly analyse our best models' results on the test set for each subtask. Figure \ref{fig:single_multi_graphs} (top) shows a comparison between our best ensemble model's predictions for Subtask 1 (Ensemble 2\textsubscript{single\_word}) and the gold answers. We observe that our model often fails to predict correctly in the range where samples have a complexity score below 0.2. We hypothesize this might be due to the skewed distribution of the golden complexity scores for each domain, as shown in Table~\ref{tab:single_scatter}.
A possible solution might be to build domain-specific models, and we plan to explore this in future efforts.

\begin{table}
\small
\centering
\begin{tabular}{@{\hskip1pt}c@{\hskip1pt}|@{\hskip3pt}c@{\hskip3pt}|@{\hskip3pt}c@{\hskip3pt}|@{\hskip3pt}c@{\hskip3pt}} \hline
\toprule
& \bf{Bible}  & \bf{Europarl} & \bf{Biomed} \\ \hline
\multicolumn{4}{c}{Sub-task 1} \\ \hline
&
\includegraphics[scale=0.1]{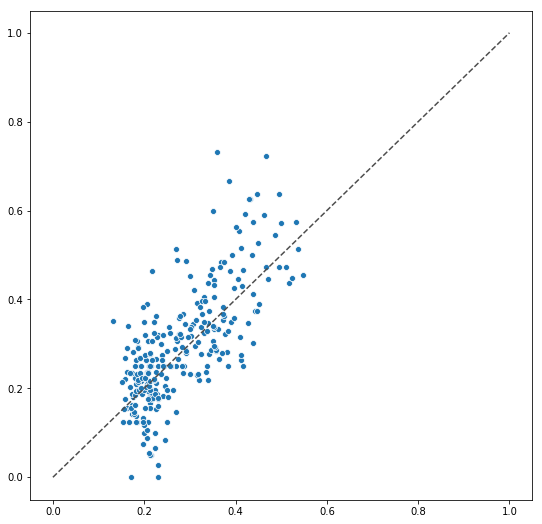} &
\includegraphics[scale=0.1]{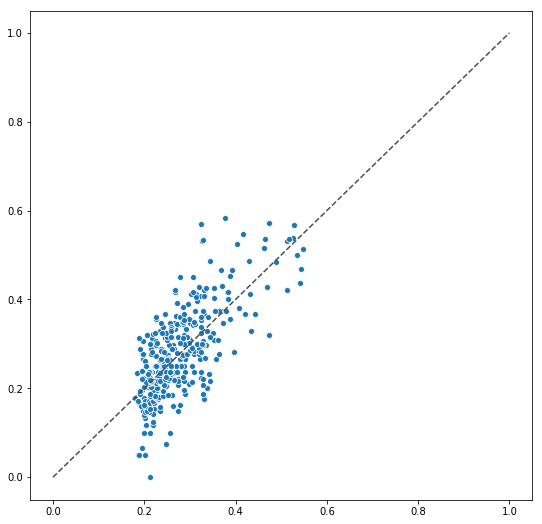} &
\includegraphics[scale=0.1]{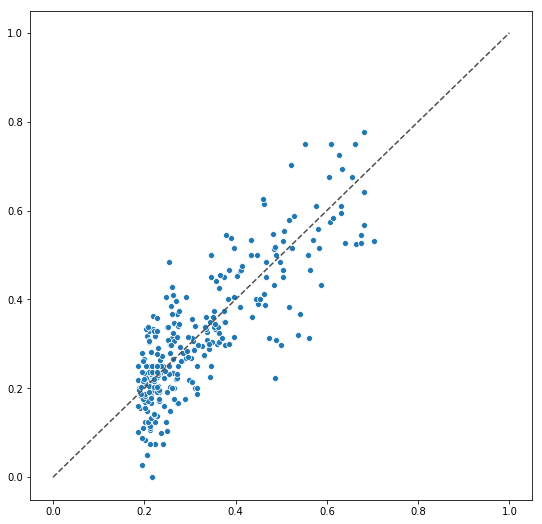} \\ \hline
\bf{MAE}     & 0.0679 & 0.0549 & 0.0638 \\ \hline
\bf{R} & 0.7329 & 0.7213 & 0.8358 \\ \hline \hline
\multicolumn{4}{c}{Sub-task 2} \\ \hline

&
\includegraphics[scale=0.1]{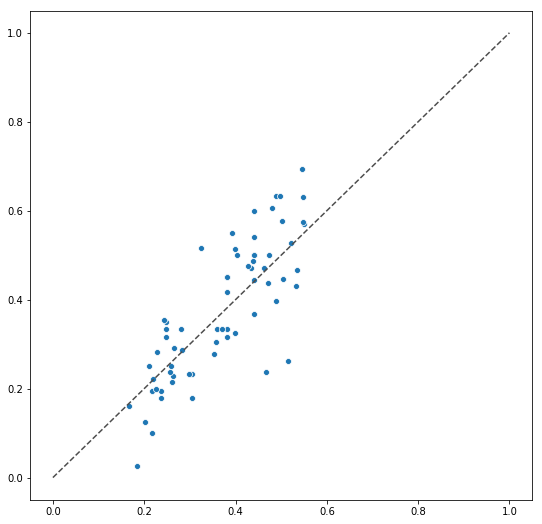} &
\includegraphics[scale=0.1]{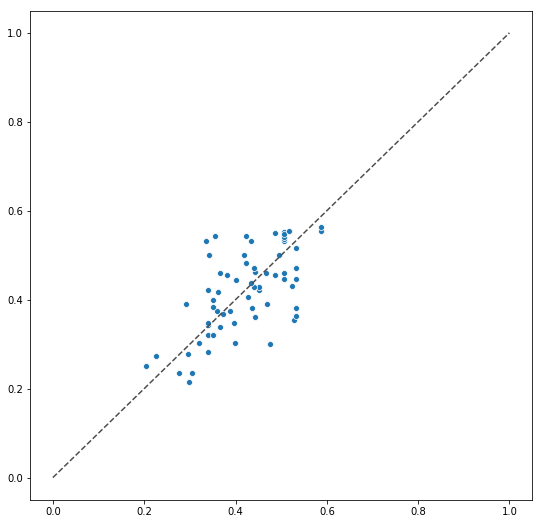} &
\includegraphics[scale=0.1]{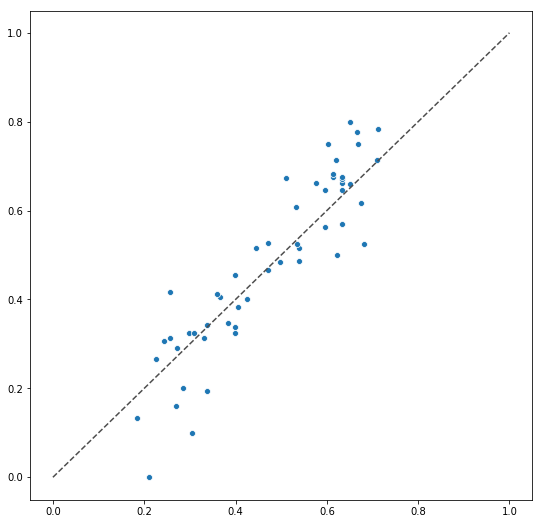} \\ \hline
\bf{MAE}     & 0.0721 & 0.0592 & 0.0667 \\ \hline
\bf{R} & 0.8114 & 0.6374 & 0.9104 \\
\bottomrule
\end{tabular}
\caption{\label{tab:single_scatter} Performance of Ensemble 2\textsubscript{single\_word} and Ensemble 1\textsubscript{multiword} models on each domain and subtask.}
\end{table}

Figure \ref{fig:single_multi_graphs} (bottom) shows a comparison between the best ensemble model's predictions for Subtask 2 (Ensemble 1\textsubscript{MWE}), and the gold answers.
Compared to Subtask 1, the data distribution of the development and test sets of Subtask 2 look more similar, hence a possible reason why the development and test set scores were closer than in Subtask 1 (the best ensemble models obtained development and test set scores of 0.8570 and 0.7772, respectively, in Subtask 1, and 0.8461 and 0.8438, respectively, in Subtask 2). Table~\ref{tab:error_samples} shows examples of successful and poor predictions made by Ensemble 2\textsubscript{single\_word} and Ensemble 1\textsubscript{MWE} models. Table~\ref{tab:single_scatter} shows how the performance of these models varies across domains. As we can see, the Biomedical domain obtained highest Pearson correlation scores on both subtasks. \\

\section{Conclusion}
In this paper, we have presented the implementation of the Ochadai-Kyoto system submitted to the SemEval-2021 Task 1. Our model ranked 7th out of 54 participating teams on Subtask 1, and 8th out of 37 teams on Subtask 2. We have proposed an ensemble model that leverages pretrained representations (i.e. BERT and RoBERTa), multi-step fine-tuning, multi-task learning and adversarial training. Additionally, we propose to enrich contextual representations by incorporating hand-crafted features during training. In future efforts, we plan to further improve our model to better handle data-constraint and domain-shift scenarios.

\bibliographystyle{acl_natbib}
\bibliography{anthology,acl2021}


\end{document}